\documentclass{isprs} 
\usepackage{subfigure}
\usepackage{setspace}
\usepackage{geometry} 
\usepackage{epstopdf}
\usepackage[labelsep=period]{caption}  
\usepackage[british]{babel} 
\usepackage[hang]{footmisc}
\usepackage{enumitem}
\usepackage{xcolor}
\usepackage{amsmath}
\usepackage{booktabs}
\usepackage{multirow}
\usepackage{amssymb}
\usepackage{graphicx}
\usepackage{float} 
\usepackage[section]{placeins}
\usepackage{stfloats}
\usepackage{acronym}
\usepackage{natbib}

\begin{document}

\emergencystretch=3em

\newacro{AuM}{Area under the Margin}
\newacro{CE}{Cross entropy}
\newacro{CL}{Confident learning}
\newacro{RS}{Remote sensing}
\newacro{SMAPE}{Symmetric mean absolute percentage error}
\newacro{TopoFilter}{Topological noise filtering}

\title{An assessment of data-centric methods for label noise identification in remote sensing data sets}
\date{}


\author{
 Felix Kröber\textsuperscript{1, 2},
 Genc Hoxha\textsuperscript{2},
 Ribana Roscher\textsuperscript{2}
}

\address{
	\textsuperscript{1 }Institute of Bio- and Geosciences, Forschungszentrum Jülich, Leo-Brandt-Straße, 52425 Jülich, Germany - f.kroeber@fz-juelich.de\\
	\textsuperscript{2 }Institute of Geodesy and Geoinformation, University of Bonn, Germany - (fkroeber, genc.hoxha, ribana.roscher)@uni-bonn.de\\
}



\abstract{
Label noise in the sense of incorrect labels is present in many real-world data sets and is known to severely limit the generalizability of deep learning models. In the field of remote sensing, however, automated treatment of label noise in data sets has received little attention to date. In particular, there is a lack of systematic analysis of the performance of data-centric methods that not only cope with label noise but also explicitly identify and isolate noisy labels. In this paper, we examine three such methods and evaluate their behavior under different label noise assumptions. To do this, we inject different types of label noise with noise levels ranging from 10 to 70\% into two benchmark data sets, followed by an analysis of how well the selected methods filter the label noise and how this affects task performances. With our analyses, we clearly prove the value of data-centric methods for both parts – label noise identification and task performance improvements. Our analyses provide insights into which method is the best choice depending on the setting and objective. Finally, we show in which areas there is still a need for research in the transfer of data-centric label noise methods to remote sensing data. As such, our work is a step forward in bridging the methodological establishment of data-centric label noise methods and their usage in practical settings in the remote sensing domain.
}

\keywords{Data-centric Learning, Label Noise, Area under the Margin, Topological Noise Filtering, Confident Learning, Remote Sensing}

\maketitle

\section{Introduction}\label{INTRO}

\begin{figure*}[!b]
\begin{center}
	\includegraphics[width=1.95\columnwidth]{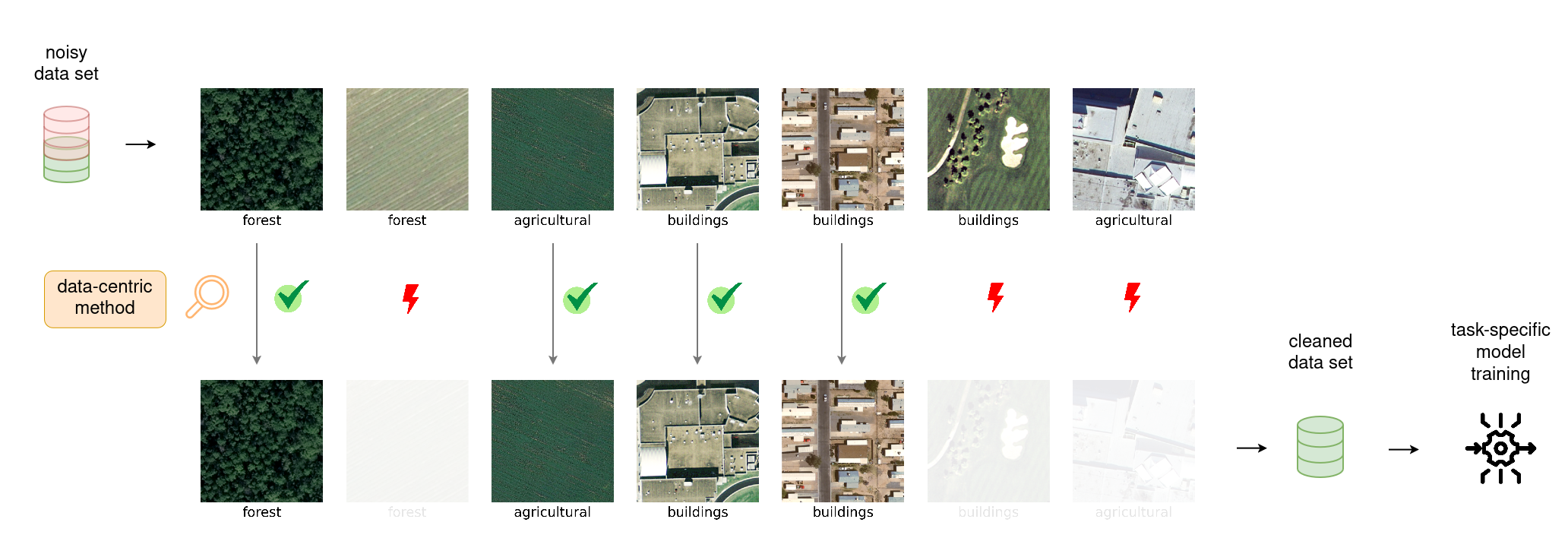}
	\caption{High-level overview of using data-centric methods to filter label noise and obtain a cleaned data set for model training.}
\label{fig:graphical_abstract}
\end{center}
\end{figure*}

Scene classification is a fundamental task in \ac{RS} aiming at assigning single labels, such as land use land cover categories, to images. Recent advances in \ac{RS} scene classification are driven by state-of-the-art deep learning methods \citep{cheng_remote_2020}. However, the effectiveness of deep neural networks heavily depends on the availability of large amounts of training data carefully annotated by human experts. The annotation process is both time-consuming and costly. To mitigate this cost, researchers often rely on crowdsourcing or publicly available thematic products such as the CORINE Land Cover map \citep{buttner_corine_2004}, GLC2000 \citep{bartholome_glc2000_2005} or GlobCover \citep{arino_global_2012}. While less costly, these labeling strategies frequently introduce noise. In the case of crowdsourcing, non-expert annotators may mislabel images due to limited domain knowledge, whereas the use of thematic products can introduce noise due to imprecise mapping or outdated information. Because over-parameterized neural networks can memorize such noisy labels \citep{zhang_understanding_2016}, performance may subsequently degrade depending on noise type and level as well as data set characteristics such as size and number of classes \citep{chen_understanding_2019,karimi_deep_2020,oyen_robustness_2022}.

To address this challenge, numerous approaches have been developed to improve model performance under noisy labels \citep{song_learning_2022}. These approaches can be broadly characterized into model-centric or data-centric strategies. Model-centric approaches focus on developing noise-robust models by introducing architectural modifications, designing noise robust loss functions, or incorporating regularization techniques. On the other hand, data-centric approaches aim to improve the quality of the training data by detecting and removing noisy samples and (re-)training the deep learning model on the cleaned data set. In the \ac{RS} domain, most of the currently used approaches are model-centric \citep{damodaran_entropic_2020,kang_robust_2020,burgert_effects_2022}. While effective, these approaches do not explicitly detect and filter out the noisy samples and thus do not yield a cleaned data set. However, explicit consideration of data quality and automated methods for determining an improved data set is highly relevant for the sustainable further development of machine learning for \ac{RS} \citep{roscher_better_2024}.

In this paper, we adopt a data-centric perspective to tackle label noise in \ac{RS} benchmark data sets (Figure~\ref{fig:graphical_abstract}).
We compare and evaluate a range of methods that detect and filter the noisy samples and (re-)train the model on the cleaned data to assess their impact on scene classification performance. Beyond potential classification improvements, the data-centric methods analyzed in this paper offer two additional advantages: (i) they have the potential to construct reusable, noise-reduced benchmark data sets, and (ii) they allow to analyze label noise properties present in existing data sets explicitly. Using a controlled experimental setup with synthetically generated noise, we present a framework to assess those benefits and insights gained by data-centric label noise identification methods, which cannot be obtained by noise robust model-centric approaches.


Our specific contributions are as follows:
\begin{enumerate}
\item To the best of our knowledge, we are the first ones to provide a benchmark of state-of-the-art data-centric methods for label noise identification on two \ac{RS} data sets.
\item Under a wide regime of noise types and levels, we decompose the performances of the methods into two sub-aspects by providing a) in-depth analyses of filtering performances patterns and b) an assessment of the impact of removing noisy labels on \ac{RS} scene classification performance.
\end{enumerate}

The remainder of this paper is organized as follows: Section~\ref{sec:related_works} briefly summarizes existing works for handling label noise. Data and methods used in our work are detailed in Section~\ref{sec:methods}. The analysis of results takes place in Section~\ref{sec:results}. Finally, Section~\ref{sec:Conclusions} concludes with a summary of the main findings. 
\sloppy

\section{Related works}\label{sec:related_works}

Methods for handling noisy labels in the computer vision community can be broadly categorized into model-centric and data-centric approaches. Model-centric approaches address label noise through: i) the development of deep learning architectures that explicitly model noise, ii) the design of noise robust loss functions and iii) the employment of regularization strategies. Architectural methods focus on estimating the latent clean label distributions and learning a noise-transition function that maps latent clean labels to the observed noisy labels. For example, \citep{sukhbaatar_training_2014} introduce an additional linear layer on top of a convolutional neural network to map clean labels to noisy labels. \citep{yao_deep_2018} propose a contrastive-additive noise network that first estimates the quality of labels and then aggregates the class predictions and noisy labels. The second category of methods focus on designing robust loss functions to alleviate the noise effects \citep{ghosh_robust_2017}. \citep{zhang_generalized_2018} introduce the generalized \ac{CE} loss function as an interpolation between the cross-entropy loss and mean absolute error to improve the robustness to noisy labels. \citep{wang_symmetric_2019} propose a symmetric \ac{CE} loss function to address the problem of overfitting of noisy labels for easy samples and the under-learning for hard samples. This is achieved by combining the standard \ac{CE} loss function with a reverse \ac{CE} term. The former provides strong optimization and convergence properties, while the latter enhances robustness to label noise. Similarly, \citep{ma_normalized_2020} propose an Active Passive Loss function that combines active losses (e.g. \ac{CE} loss) to encourage optimization and convergence properties with passive losses (e.g. Mean Absolute Error) to improve robustness to noisy labels. The third category of model-centric approaches uses regularization strategies to prevent overfitting of the models to noisy labels. Examples of such strategies include Early regularization \citep{liu_early-learning_2020,xia_robust_2020}, Co-teaching \citep{han_co-teaching_2018} or Mixup \citep{zhang_mixup_2017}. Early regularization techniques introduce a term into the loss that encourages consistency with early-stage class predictions, and thus preventing the model from memorizing noisy-labels. Co-teaching consists of training two networks simultaneously. Each network acts as a supervisor for the other by selecting samples characterized by small loss, thus reducing the influence of noise-labels. Mixup on the other hand is based on data augmentation and creates virtual training samples utilizing linear interpolation of both input features and corresponding labels.
  
Data-centric approaches instead tackle the problem of learning under noisy labels by explicitly detecting and filtering the noisy samples in order to (re-)train the model on the cleaned data. To this end, data-centric modeling for label noise identification usually first starts with the training of a neural network on the noisy data set. Since the data-centric methods are mostly model-agnostic, any deep neural network architecture can be used. Various parts of the information contained in or issued from the training process are subsequently leveraged to identify noisy labels and separate them from the clean data. A first group of methods exploits the learned feature space, assuming that noisy labels exhibit distinct geometric or topological neighborhood structures \citep{wu_topological_2020, wu_ngc_2021}. A second line of work relies on training dynamics, leveraging the evolution of metrics over several training epochs to distinguish noisy samples from reliable ones \citep{huang_o2u-net_2019, nguyen_self_2019, pleiss_identifying_2020}. A third group focuses on the final prediction layer of a preliminary trained model, using the logits or predicted class probabilities to identify inconsistent or low-confidence samples \citep{chen_understanding_2019,bahri_deep_2020,northcutt_confident_2021}. Dependent on the specific data-centric method, the noisy label filtering can either be done in an online-fashion during the initial training of the network or at the end of the training process. In the former case, both the information on the filtered noisy data set and the task performance for training on cleaned data can be obtained from the single model training run. In cases where the separation of clean and noisy data requires the training on the noisy data to be completed, a second stage with the re-training of the network on cleaned data is necessary to obtain the task performance and assess it’s benefit compared to training on noisy labels.

In \ac{RS}, most works follow a model-centric approach. As an example, \citep{kang_robust_2020} integrate a downweighting factor into the softmax loss to reduce the contributions of the noisy images when learning class prototypes. \citep{sumbul_generative_2023} combine a supervised variational autoencoder with a discriminative network to identify noisy labels and reduce their influence during training. A collaborative learning framework building upon co-training is proposed in \citep{aksoy_multi-label_2022} in the context of multilabel classification. Similarly, \citep{otsu_robust_2025} propose noise robust co-training with automatic filtering for \ac{RS} image segmentation. These methods effectively adapt model-centric approaches from the computer vision domain to the peculiarity of \ac{RS} images and tasks. However, they focus on coping with noisy labels rather than explicitly separating clean and noisy samples. A corresponding consideration of the added value of data-centric methods in this regard is still lacking in the \ac{RS} domain.   

\section{Data \& Methods}\label{sec:methods}

\subsection{Data sets \& experimental setup}

With UCMerced \citep{yang_bag--visual-words_2010} and EuroSAT \citep{helber_eurosat_2019}, we select two well-studied benchmark data sets to assess the capabilities of data-centric methods to handle added label noise. Both data sets are known to be high-quality data sets, as they have been manually labeled and reviewed multiple times, which means they can be assumed to be nearly free of labeling errors. This enables artificial label noise to be induced in a controlled manner according to given probabilistic noise transition matrices. We use the following noise models that are commonly used in existing studies: 

\begin{itemize}
    \item uniform noise: class-independent noise where every true label $y*$ has the same probability of being flipped to any other class $\widetilde{y}$.
    \item asymmetric noise: class-dependent noise where some classes are more prone to mislabeling than others. Each class y has its own noise level, sampled from a normal distribution around the target. The noise is distributed to a subset of other classes determined by a random sparsity mask, controlling how many off-diagonal elements receive nonzero probabilities where we consider two different sparsity levels of 0.25 and 0.75.
\end{itemize}

\begin{figure}[ht!]
\begin{center}
		\includegraphics[width=0.65\columnwidth]{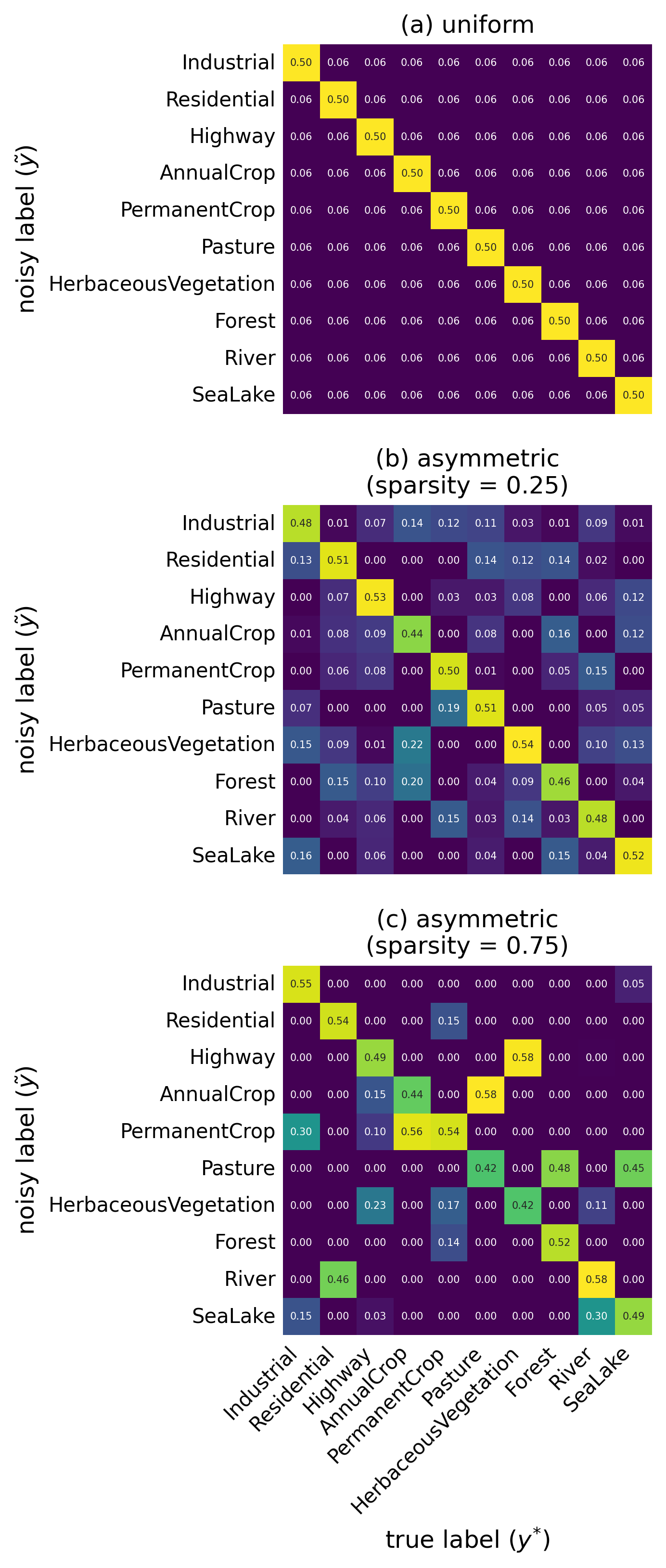}
	\caption{Exemplary noise transition matrices for EuroSAT at an overall noise level of 50\%.}
\label{fig:noise}
\end{center}
\end{figure}

Compared to uniform noise, asymmetric sparsity-controlled noise is harder to cope with \citep{northcutt_confident_2021}. It represents structured noise more closely reflecting real-world conditions, e.g. the noise transition probability between two semantically similar classes differing from the noise transition probability between two clearly separable classes.  

Exemplary resulting noise transition matrices are represented in Figure~\ref{fig:noise}. The overall noise level is specified by the trace of the matrices. We assess a total of four noise levels ranging from 10\% to 70\%. Sampling from the noise matrices, faulty train data sets are created that are subsequently used for the data-centric workflow as described in the Section~\ref{DCAI}. Noise is only applied to the training data sets, which are sampled to be 80\% of the total data set. Validation and test sets (10\% each) are left untouched to evaluate the task performance of the models on the flawless data. For computational reasons, the definition of train, validation and test splits for EuroSAT relies on a subsampled version of original data set containing 2.7K images. For all training runs carried out subsequently, we rely on a ResNet-18 \citep{he_deep_2016} trained for 50 epochs (UCMerced) and 100 epochs (EuroSAT), respectively. Based on preliminary experiments assessing the performance on the validation set, a learning rate of 1e-5 and batch size of 32 are set as hyperparameters. We rely on the standard Adam optimizer \citep{kingma_adam_2014} and use a cross-entropy loss. To have statistically meaningful results, we run each of the experimental configurations consisting of noise matrix and network initialization three times with differing seeds and aggregate the results across those.

\FloatBarrier
\subsection{Data-centric methods}\label{DCAI}

With the specific selection of three data-centric methods for label noise identification, we attempt to cover existing methodological branches as comprehensively as possible. Each of the methods uses different pieces of information from the original training process on noisy data targeting the feature space structure, logit evolution or predicted probabilities, respectively.

\begin{enumerate}[label=(\alph*)]
    \item \ac{TopoFilter} \citep{wu_topological_2020} treats the identification of clean samples as a topological problem. It assumes that clean samples form dense clusters in the feature space, whereas noisy samples appear as outliers or peripheral points. The method alternates between training a classifier and selecting clean data. At each alternation, it builds a k-nearest neighbor graph from the latent representations and extracts, for each class, the largest connected component. To further improve purity, \ac{TopoFilter} performs $\zeta$-filtering, removing points within the connected components whose local neighborhoods contain too many samples of different labels. The method iteratively refines the network representation using only the selected clean data, progressively improving both representation quality and data purity. For all of our experiments, we first run the topological filter at the 10\textsuperscript{th} epoch of training, repeating it every 5 epochs onwards from there.
    \item \ac{AuM} \citep{pleiss_identifying_2020} identifies mislabeled samples by analyzing the evolution of the model’s margins which are the differences between the logit of the assigned class and that of the most competitive alternative class. For each sample, the \ac{AuM} is computed as the average margin over training epochs. Clean examples, which reinforce generalizable features, tend to have high \ac{AuM} values, while mislabeled ones, which produce conflicting gradient signals, show persistently low or negative \acp{AuM}. To automatically determine below which \ac{AuM} threshold samples should be identified as mislabeled, some examples are deliberately assigned to an introduced extra class. Mimicking mislabeled behavior they allow to define an empirical \ac{AuM} threshold. After computing \acp{AuM} and applying the learned threshold, low \ac{AuM} samples are removed, and the model is re-trained on the cleaned data.
    \item \ac{CL} \citep{northcutt_confident_2021} aims to detect and correct label errors by estimating the joint distribution between noisy labels and their unknown true labels. It first uses a trained model`s predicted class probabilities, obtained from cross-validation, to compute a confident joint matrix. This records how often examples labeled as class $i$ are confidently predicted as class $j$, where “confident” translates to a the predicted class probability $p(i)$ being above a per-class threshold derived as the average predicted probability of this class. From this joint matrix, \ac{CL} estimates how many samples are mislabeled between each class pair, prunes them and allows to re-train the model on the cleaned data set.
\end{enumerate}

\subsection{Evaluation}

During evaluation, we first investigate the capabilities of the methods to identify noisy train labels correctly (Section~\ref{sec:noise_ident}).  To this end, we use a total of five metrics starting with the calculation of (1) precision, (2) recall and (3) remaining noise level (Section~\ref{sec:micro}). Precision and recall quantify the proportion of correctly detected noisy labels among all predicted noisy labels and among all true noisy labels, respectively. The remaining noise corresponds to the fraction of mislabeled samples that remain in the data set after cleaning. Furthermore, we evaluate (4) \textit{$\Delta$} noise level, and (5) \ac{SMAPE} to quantify the biases of the methods in estimating overall noise levels across the full data set (Section~\ref{sec:macro}). Here, $\Delta$ noise level denotes the difference between the predicted and true noise proportions, i.e.\ the deviation between the estimated and actual fraction of noisy labels. \ac{SMAPE} is calculated as $$2 \cdot \frac{|\hat{p}_{\text{noise}} - p_{\text{noise}}|}{|\hat{p}_{\text{noise}}| + |p_{\text{noise}}|} \times 100,$$ where $\hat{p}_{\text{noise}}$ and $p_{\text{noise}}$ are the predicted and true noise levels, respectively, providing a symmetric percentage error that accounts for over- and underestimation equally. 

Subsequently, we use the clean test set to assess changes in task performances when comparing models trained on cleaned data to the baseline when training on noisy data (Section~\ref{sec:Taskper}). To quantify how well each method improves the task performance after the cleaning process, we primarily report and compare overall accuracies.

Finally, we examine the effect of the noise filtering capabilities of the methods on the task performance changes (Section~\ref{sec:Synergy}). In order to comprehensively and adequately map complex correlations - likely to be non-linear, multivariate and only valid for subgroups - we perform a manual cluster analysis. Specifically, we analyze clusters of constellations consisting of data sets, noise types, noise levels and data-centric methods, which exhibit largely homogeneous properties in terms of filtering and subsequent task performances. The basis for this cluster analysis is a dimension-reduced visualization of the 5-dimensional space spanned by the aforementioned metrics used to assess the filtering properties. For dimension reduction, we use the non-linear uniform manifold approximation and projection method (UMAP) \citep{mcinnes_umap_2018}.

\section{Results}\label{sec:results}
\subsection{Noisy label identification}\label{sec:noise_ident}

\subsubsection{Microscopic view}\label{sec:micro}
\begin{figure*}[ht!]
\begin{center}
	\includegraphics[width=2.0\columnwidth]{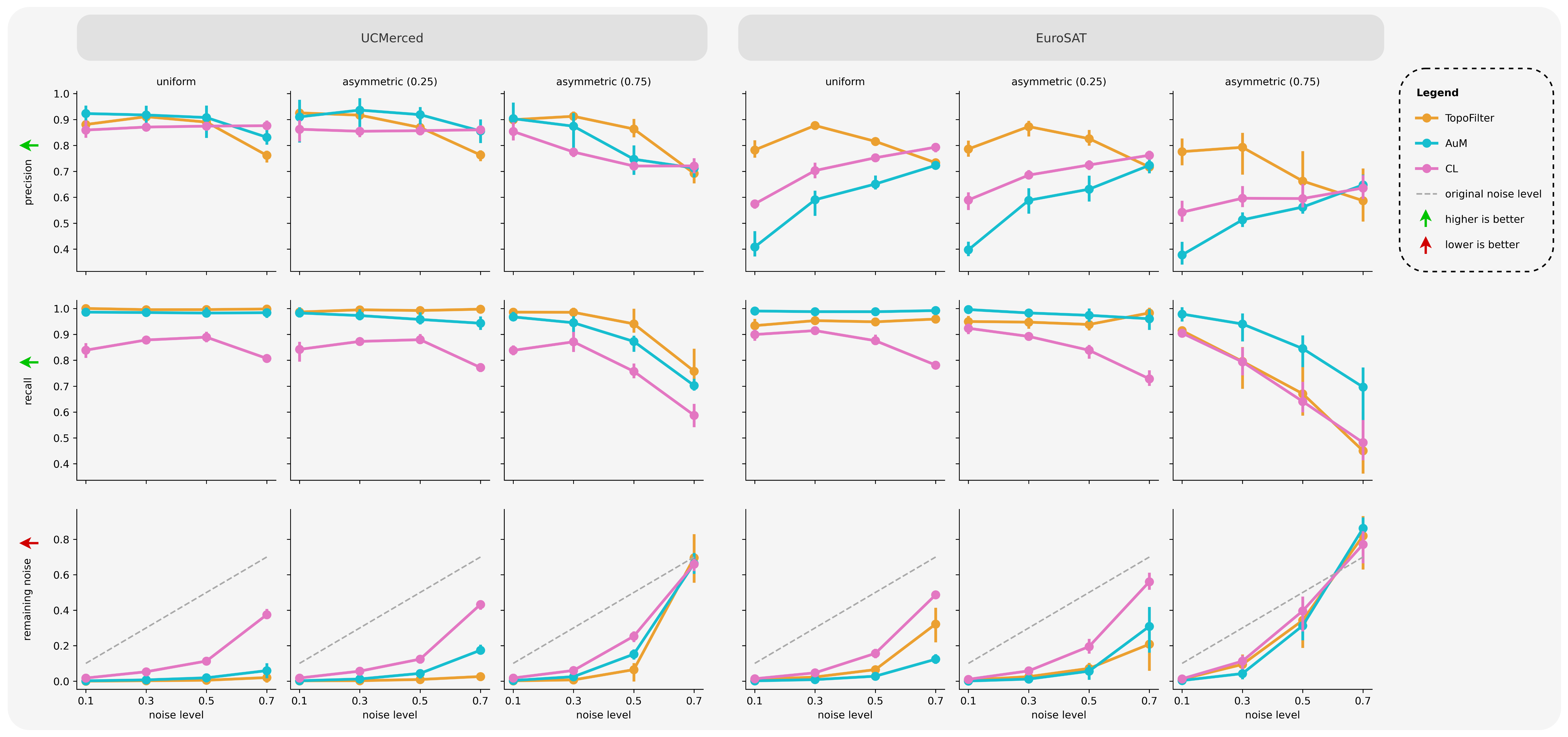}
	\caption{Statistics on label noise identification across settings.}
\label{fig:results_aI}
\end{center}
\end{figure*}

From the analysis of precision and recall in noise detection (Figure~\ref{fig:results_aI}), it is first of all evident that all data-centric methods are capable of detection label noise well with precision and recall values between 0.4 and 1 depending on the data set, noise type and noise level. Aggregating precision and recall into F1 scores, even for the most difficult setting – asymmetric noise with high sparsity applied to EuroSAT with 70\% noise – the worst performing method \ac{TopoFilter} still reaches a decent F1 score of $\geq$ 0.5. In most cases, the precise and comprehensive detection of label noise translates into a clearly visible reduction in label noise in the cleaned data set compared to the initial noise level. Ideally, the remaining noise level as shown in Figure~\ref{fig:results_aI} should remain close to zero and clearly below the dashed reference line representing the original noise level. The consistent gap between the remaining noise and the reference line across most conditions confirms that all methods substantially reduce the overall label noise. Only at noise levels where the label noise dominates the signal, methods may no longer able to reduce the noise level. This indicates limitations in fully identifying corrupted labels under severe contamination. A more detailed breakdown of this overall summary reveals systematic and significant differences in noise detection capabilities for the different methods across all of the considered factors – noise types, noise levels and data sets. 

Starting with noise types, precision and recall values for uniform noises are highest meeting the expectation that uniform noise is easy to be handled. For asymmetric noise with low sparsity very similar results to uniform noise can be observed with almost no changes in the metrics. This may be explained by the effective signal-to-noise ratio as the ratio between the frequency of correct labels for a given class and the frequency of the label with which this class is most frequently confused. In the low-sparsity asymmetric case, this ratio remains high because the noise is still spread across several classes, preserving a relatively strong signal. In contrast, in the high-sparsity asymmetric scenario, the noise becomes concentrated in only a few confusions, substantially lowering the effective signal-to-noise ratio and resulting in almost continuously reduced precision and recall. This aligns with high sparsity asymmetric noise cases representing the only ones where the methods do not reduce the noise level when cleaning the data under high noise levels.  

As for noise levels, precision and recall figures tend to decrease with increasing noise levels reflecting the growing difficulty of correctly distinguishing corrupted from clean labels in highly noisy conditions. Nevertheless, the decrease is not uniform across data sets. On EuroSAT, precision slightly improves at higher noise levels. Such non-decreasing precision or recall trends can be explained by the fact that, when models correctly estimate the overall fraction of noisy samples (Section~\ref{sec:macro}), even a random identification of noisy labels will yield increasing precision and recall values for increasing noise levels. Complementary, we expect an unskilled random guesser of label noise to obtain precision and recall values close to zero at low noise levels. This emphasizes the significance of the almost perfect noise detection figures observed for the analyzed methods in such low noise regimes. Even in the worst-case scenario, the noise level after noise removal is only 1.76\% with 10\% injected noise and 11.28\% with 30\% noise. Taken together, the results show that all methods perform well in low-noise regimes — where precision, recall, and remaining noise approach their ideal limits — and gradually deteriorate as the noise level increases.

In terms of data sets, there are clearly visible difference in patterns and accuracies between both UCMerced and EuroSAT. Broadly speaking, EuroSAT proves to be more difficult for noise removal. The remaining noise level after noise removal is systematically higher than with UCMerced. While the methods applied to UCMerced are still able to reduce the noise level even with 70\% noise and asymmetric noise, this is no longer the case with EuroSAT. At lower noise levels, both data sets show significant differences in terms of the precision values achieved. For UCMerced in particular, we achieve high precision values at low noise levels, while for EuroSAT we observe lower precision values with all methods. The ranking of the methods among themselves also changes when comparing the two data sets. While \ac{AuM} often achieves the best precision on UCMerced, the method achieves the worst values on EuroSAT. In terms of recall values, \ac{TopoFilter} outperforms \ac{AuM} for UCMerced, while the opposite is true for EuroSAT. These significant differences are contradicting the expectation that both data sets behave similarly due to their similarity in terms of being class-balanced and having a low to medium number of classes easy to be differentiated. Still, EuroSAT is generally considered somewhat more difficult to classify than UCMerced. EuroSAT is less object-focused, and confusion between classes (e.g., pasture and herbaceous vegetation) may occur more easily. Accordingly, the initial noise level, which is assumed to be negligible, could already be slightly higher without injected noise for EuroSAT, thus providing a plausible explanation for some of the observed data set-specific differences.

\begin{table}[H]
\centering
\captionsetup{justification=justified,singlelinecheck=false}
\caption{Summary of times a given data-centric method performs best or worst across all settings (4 noise levels, 3 noise types \& 2 data sets) specified for the noise identification metrics.}
\resizebox{\columnwidth}{!}{%
\begin{tabular}{llccc}
\toprule
 &  & \textbf{TopoFilter} & \textbf{AuM} & \textbf{CL} \\
\midrule
\multirow{3}{*}{\textbf{Wins \textcolor{green!60!black}{$\uparrow$}}} 
 & precision              & \textbf{12} & 7  & 5  \\
 & recall                 & \textbf{13} & 11 & 0  \\
 & remaining noise         & \textbf{12} & 10 & 2  \\
\midrule
\multirow{3}{*}{\textbf{Losses \textcolor{red!70!black}{$\uparrow$}}} 
 & precision              & \textbf{5}  & 10 & 9  \\
 & recall                 & 1  & \textbf{0} & 23 \\
 & remaining noise         & \textbf{1}  & \textbf{1} & 22 \\
\bottomrule
\end{tabular}%
}
\label{tab:results_aI}
\end{table}

Generalisations that can be made across data sets, noise types and noise levels include that \ac{AuM} and \ac{TopoFilter} seem to prioritize recall over precision, whereas for \ac{CL} both figures are more balanced. Overall \ac{TopoFilter} performs best in most cases. According to Table~\ref{tab:results_aI}, \ac{TopoFilter} wins the method comparison 12 times for precision and the remaining noise level and 13 times for recall. It is also apparent that \ac{CL} is inferior to the other methods, especially in terms of recall where it systematically ranks below the other methods loosing 23 out of 24 times. The same applies to the remaining noise level where \ac{CL} looses 22 times. \ac{CL} outperforms other methods only in high noise, asymmetric noise scenarios.

\subsubsection{Macroscopic view}\label{sec:macro}

\begin{figure*}[ht!]
\begin{center}
	\includegraphics[width=2.0\columnwidth]{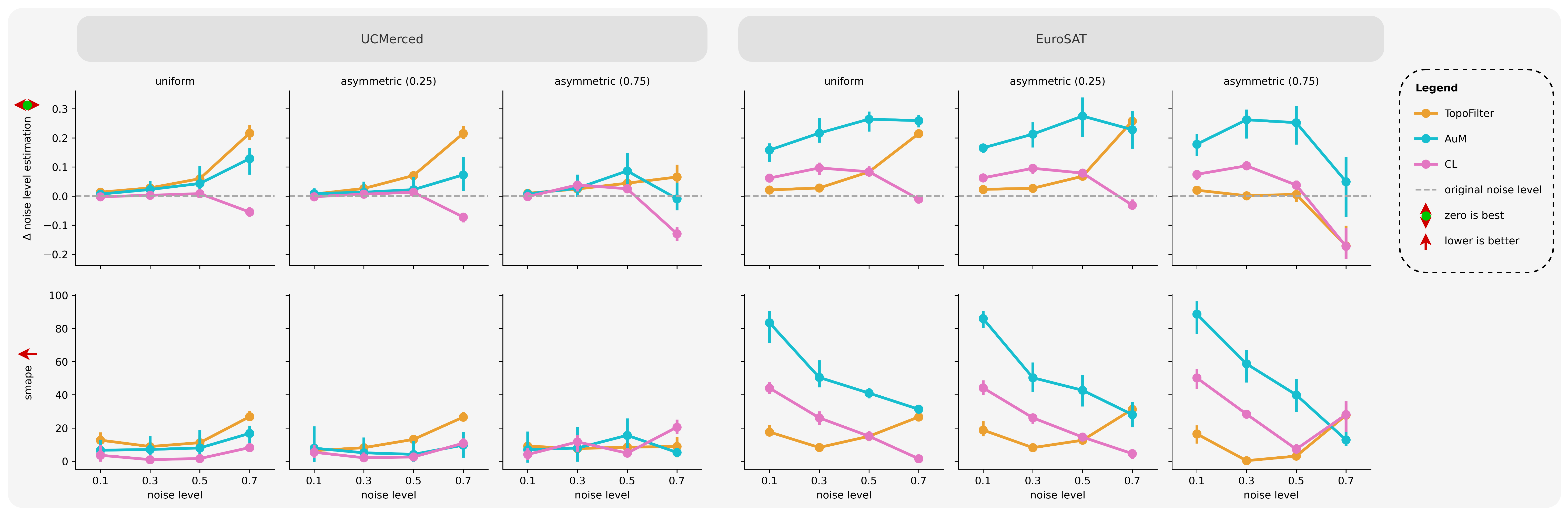}
	\caption{Statistics on data set-wide label noise level estimation across settings.}
\label{fig:results_aII}
\end{center}
\end{figure*}

Aggregating the identified individual flaws into a predicted overall noise level, the accuracy of this prediction in relation to the actual controlled noise can be analyzed. In line with the described precision and recall values, Figure~\ref{fig:results_aII} demonstrates the fundamentally good ability of the data-centric methods to estimate the overall noise level across noise types, levels and data sets. For low to medium noise levels ($\leq$ 50\%), at least one method exists that estimates the given noise level with an absolute error of less than 10 percentage points. Expressed as a relative error quantified via \ac{SMAPE}, the deviations here are less than 20\% of the target level. Correlating the estimated noise level with the target noise level yields correlation coefficients of $r \geq 0.9$ for all methods and noise types.  

A detailed analysis reveals some similar patterns to those seen in Subsection \ref{sec:micro}, especially with regard to the differences between the data sets. For EuroSAT, the poorer accuracies in label noise detection correlate, for example, with a higher bias of the methods with regard to determining the overall noise level. The average relative error across noise levels is 6.4\% (CL), 8.4\% (AuM) and 12.34\% (TopoFilter) for UCMerced. For EuroSAT, these values are higher by a factor of 1.25 – 6.1 with 24.2\% (CL), 51.1\% (AuM), and 15.5\% (TopoFilter), respectively. This variability in results between data sets is more pronounced than the influence of noise levels and noise types. The noise type in particular has almost no influence on the accuracies achieved. Unlike the estimation of precision and recall values, asymmetric noise with high sparsity shows only slightly altered patterns compared to uniform or low sparsity noise. The estimation accuracy of the overall noise level appears to be insensitive to the specific noise type for all methods. For noise level, the expected correlation applies, namely that increasing noise levels tend to lead to a higher absolute error, but this does not necessarily have to be accompanied by a higher relative error. For UCMerced, the relative error across noise levels is reasonably stable, and for EuroSAT, it even decreases for \ac{AuM} and \ac{CL}.

\begin{table}[H]
\centering
\captionsetup{justification=justified,singlelinecheck=false}
\caption{Summary of times a given data-centric method performs best or worst across all settings (4 noise levels, 3 noise types \& 2 data sets) specified for the noise level estimation metrics.}
\resizebox{0.9\columnwidth}{!}{%
\begin{tabular}{llccc}
\toprule
 &  & \textbf{TopoFilter} & \textbf{AuM} & \textbf{CL} \\
\midrule
\multirow{2}{*}{\textbf{Wins \textcolor{green!60!black}{$\uparrow$}}} 
 & $\Delta$ noise level  & 10 & 2  & \textbf{12} \\
 & SMAPE                  & 10 & 3  & \textbf{11} \\
\midrule
\multirow{2}{*}{\textbf{Losses \textcolor{red!70!black}{$\uparrow$}}} 
 & $\Delta$ noise level  & 9  & 12 & \textbf{3}  \\
 & SMAPE                  & 9  & 12 & \textbf{3}  \\
\bottomrule
\end{tabular}%
}
\label{tab:results_aII}
\end{table}

Comparing the different data-centric methods, the following can be generalized: Despite its previously noted modest performance in terms of recall and precision, \ac{CL} performs best in estimating the overall noise level (Table~\ref{tab:results_aII}). Compared to \ac{TopoFilter} and \ac{AuM}, it usually has a lower bias and is less prone to overestimating the noise level at low noise levels. At high noise levels, \ac{CL} tends to underestimate the actual noise level. \ac{AuM} delivers rather mediocre results in quantifying the noise level, but these are also subject to considerable fluctuations relative to the other methods, depending on the random initialization of the noise transition matrix or the training. With the exception of very high noise levels, \ac{TopoFilter} performs similarly well to CL.

\subsection{Task performance}\label{sec:Taskper}

\begin{figure*}[ht!]
\begin{center}
	\includegraphics[width=2.0\columnwidth]{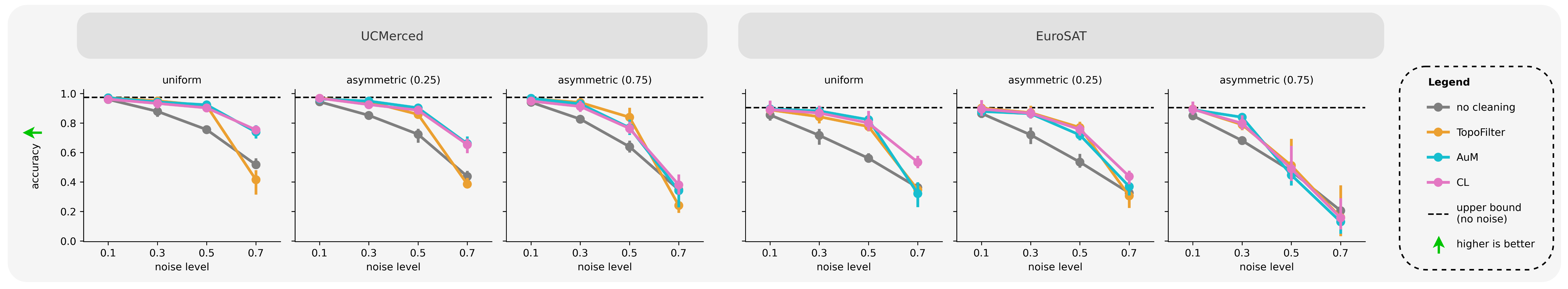}
	\caption{Scene classification test accuracies when training on cleaned and noisy data.}
\label{fig:results_b}
\end{center}
\end{figure*}

The results of the evaluation of the models trained on cleaned and uncleaned data are summarized in Figure \ref{fig:results_b}. Training the models without injected noise achieves very good values for both data sets with accuracies $\geq0.9$.
This is in line with expectations for the data sets, which are easily classifiable benchmarks with high-quality label information, meaning that good training results translate into correspondingly high test accuracies. Without the application of one of the data-centric methods examined, accuracy decreases continuously as noise increases. At a noise level of 10\%, the performance gap compared to the results without injected label noise is at least -1.4\% (UCMerced, uniform noise) and ranges up to -5.5\% (EuroSAT, asymmetric high sparsity noise). Up to a noise level of 70\%, the performance gap grows to at least -45.7\% and peaks at -69.9\%. Although the negative effects of massive label noise could potentially be mitigated by increasing the data set size \citep{rolnick_deep_2018}, in our setup, with the data set size remaining constant, the detrimental effect of label noise is clearly apparent. Without label noise exclusion, the trained models generalize significantly worse, and this applies to all noise types across both data sets.
 
The application of data-centric methods for label noise removal clearly shows positive effects on the task accuracies achieved, as indicated by the improvements in Figure~\ref{fig:results_b}. This applies in principle to all of the methods tested, regardless of minor differences between methods and data sets. Starting with low noise levels of 10\%, the task performance is above the respective plain baseline without noise removal, with values ranging from a minimum of 0.0\% (CL, UCMerced, uniform noise) to a maximum of 4.5\% (TopoFilter, EuroSAT, asymmetric high sparsity noise). On average across methods, noise types and data sets, accuracy improves by an absolute 2.6\%, thus compensating for a large part of the performance gap of -3.7\% caused by the injected noise when no cleaning is conducted. This also applies to medium noise levels of 30\% and 50\%, where an average improvement of 11.2\% and 15.5\% respectively is achieved compared to training without noise removal. Only at very high noise levels of 70\% the limitations of the methods become apparent. Here, an average improvement of only 4.1\% is achieved, and in the case of asymmetric high sparsity noise in particular, there is also a deterioration in task performance.

A comparison of the methods (Table~\ref{tab:results_b}) shows that \ac{TopoFilter} performs best in 11 out of 24 cases. At the same time, in 9 out of 24 cases, it is worse than both other methods. Contrary, \ac{CL} only performs best in 4 out of 24 cases, but also performs worst less often (7 out of 24). It can therefore be stated that \ac{TopoFilter} is the most promising choice in terms of task performance in many cases, but also the most risky. \ac{CL}, on the other hand, is a risk-averse choice when conditions such as noise level and noise type are unknown. Setting-specific differences between the methods are particularly evident in relation to noise level: \ac{TopoFilter} loses performance particularly at high noise levels, whereas \ac{CL} outperforms the other methods in this scenario.

\begin{table}[H]
\centering
\captionsetup{justification=justified,singlelinecheck=false}
\caption{Summary of times a given data-centric method performs best or worst across all settings (4 noise levels, 3 noise types \& 2 data sets) specified for the task performance metric.}
\scalebox{1}{%
\begin{tabular}{llccc}
\toprule
 &  & \textbf{TopoFilter} & \textbf{AuM} & \textbf{CL} \\
\midrule
\multirow{1}{*}{\textbf{Wins \textcolor{green!60!black}{$\uparrow$}}} 
 & accuracy               & \textbf{11} & 9  & 4  \\
\midrule
\multirow{1}{*}{\textbf{Losses \textcolor{red!70!black}{$\uparrow$}}} 
 & accuracy               & \textbf{9}  & 8  & 7  \\
\bottomrule
\end{tabular}%
}
\label{tab:results_b}
\end{table}

Surprisingly the methods behave relatively similarly across the two data sets. Despite strong data set-specific differences in the filtering performance of the methods (Section~\ref{sec:noise_ident}), the task performance for both data sets shows similar patterns and intensities of improvement in the values achieved by the data-centric methods. A possible connection here could be that the task performance of the methods is less dependent on the absolute precision and recall of noise detection and more on the question of whether they effectively reduce the noise level through their filtering (Figure~\ref{fig:results_aI}). Thus, the cases of deterioration in accuracy in Figure~\ref{fig:results_b} partially coincide with the cases in which the noise level has not been reduced by filtering. However, a closer look also shows that the noise-reducing property may represent at most a necessary but not sufficient condition for a positive effect on task performance. For example, \ac{TopoFilter} shows an effective reduction in noise according to Figure~\ref{fig:results_aI} under high levels of uniform noise, yet there is a deterioration in task performance. How exactly the various factors interact is therefore not trivial to analyze. The following Subsection provides a closer combined examination of input variables and filtering properties in relation to the task performances achieved.

\subsection{Synergy}\label{sec:Synergy}

\begin{figure*}[ht!]
\begin{center}
	\includegraphics[width=1.8\columnwidth]{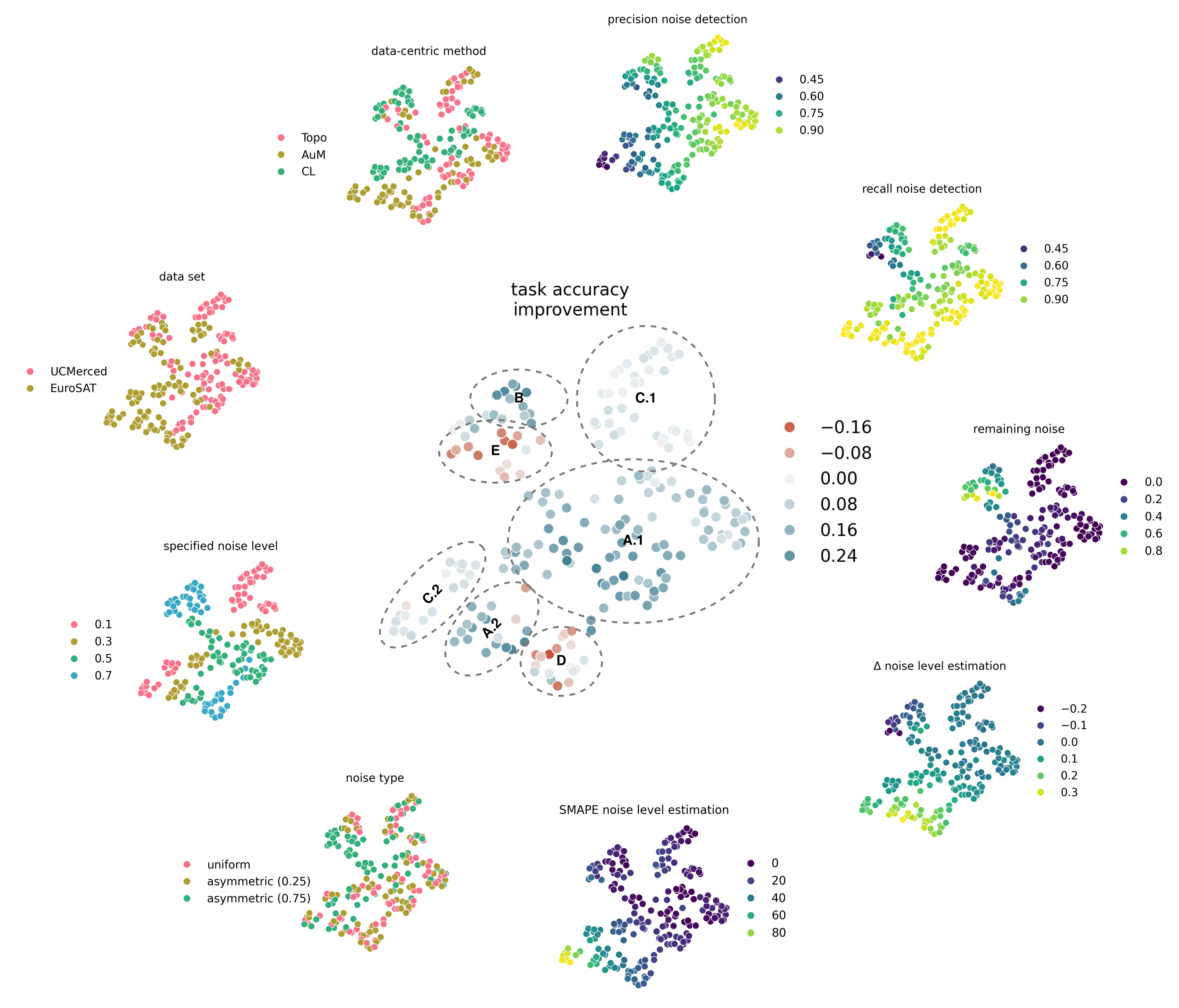}
    \captionsetup{justification=justified,singlelinecheck=false}
	\caption{UMAP-reduced representation of the label noise constellations examined. Input variables are shown in the left semicircle, the calculated label noise filtering properties in the right semicircle, and the final effect on task accuracies in the center. Task accuracy improvements are measured against the baseline of training on uncleaned data depicted as "plain" in Figure \ref{fig:results_b}. Clusters with homogeneous behavior in terms of filtering properties and task accuracy improvements are marked with dashed lines (A-E).}
\label{fig:results_c}
\end{center}
\end{figure*}

In order to systematically analyze the relationships between filtering properties and task accuracy effects of data-centric methods, clusters of constellations of methods, data sets, noise levels and noise types can be identified that behave similarly. Figure~\ref{fig:results_c} shows a total of five such clusters (A-E), which can be characterized as follows:

\begin{enumerate}[label=\Alph*.]
    \item This largest cluster covers almost all cases with medium to high noise levels of 30\% and 50\% respectively. With the exception of a few constellations that fall into cluster~E, this cluster contains all combinations of the other variables (data-centric method, data set and noise type). The influence of these variables can therefore be classified as rather irrelevant in the overall view for these noise levels. All methods behave similarly with regard to label noise identification and task accuracy changes, largely independently of noise type and data set. With regard to label filtering properties, noise is accurately detected with medium to high precision and high recall values. The remaining noise level after removal of labels identified as noise is consistently low. The error in label noise identification can vary in absolute and relative terms in parts of this cluster, as illustrated by the difference between A.1 and A.2. However, this does not change the effect of good noise detection in terms of task performance after data cleaning, which is consistently positive with values mostly well above 10\%. This cluster therefore represents the majority of cases in which the application of label noise methods across different constellations delivers generally good results.
    \item The second, significantly smaller cluster comprises approximately half of the cases characterised by very high noise levels of 70\%. For both data sets, this primarily includes the subset of uniform noises and low sparsity asymmetric noises. In terms of methods, \ac{CL} is particularly included here, with hardly any \ac{AuM} and \ac{TopoFilter} cases. This constellation behaves homogeneously with regard to label noise filtering in that noise is detected with acceptable precision and recall values, thereby reducing the effective label noise level, even though the remaining noise level is still significant based on the 70\% level. Noise level estimation works well with very low relative errors and low absolute biases. The effect on task performance is a significant increase in accuracy. This cluster thus includes cases in which, despite very high initial noise, \ac{CL} proves to be one effective data-centric method.
    \item The third cluster comprises two subclusters (C.1 and C.2) that are separated in the UMAP-reduced space. In their entirety, they cover all cases of low initial noise (10\%). Accordingly, all other constellations of method, data set and noise type are included. In terms of noise detection, the detection of noise is consistently very comprehensive, albeit not particularly precise in some cases (C.2). The less precise detection of label noise is attributable to the EuroSAT-specific cases described and correlates with some very high relative errors in noise level estimation. However, due to the comprehensive noise detection, the remaining noise level is very low in all cases (C.1 and C.2) with values close to zero. In terms of its effect on task performance, this is reflected in slightly positive changes when training on the cleaned data. Given the low initial noise level, the task improvements in absolute values are correspondingly small but relatively consistent. In summary, this cluster includes the low-noise cases in which the data-centric methods work reliably with a visible, albeit not drastic, effect on task performance.
    \item The fourth cluster comprises some of the cases of very high label noise (70\%) and, similar to cluster B, those cases that are characterized by uniform or low sparsity asymmetric noise. Complementary to cluster B, the cases of \ac{AuM} and \ac{TopoFilter} application are included here and \ac{CL} is excluded. Unlike in cluster B, noise detection is performed with very high recall, correlating with a significantly reduced remaining noise level after data cleansing. At the same time, there is a systematic overestimation of the noise proportion, in some cases more than 20\% above the specified 70\% of the data. With remaining data set sizes of less than 10\% of the original training set in some cases, this has a negative effect on task performance, as the models tend to overfit and no longer generalize. This cluster thus covers the cases described in which \ac{TopoFilter} in particular is limited in its performance at very high noise levels.
    \item This last cluster now covers the rest of the cases of very high label noise, complementing clusters B and D. What is unique here is that all cases of asymmetric noise with high sparsity are now included. Noise detection works here with moderate precision and recall values only to a limited extent. The output noise level is estimated with good accuracy, both absolutely and relatively, but noise detection is too inaccurate to bring about an effective reduction in the noise level by cleaning the data. This is reflected in predominantly, and in some cases significantly, poorer task performance. This cluster thus subsumes cases in which data-centric methods are limited in terms of both noise detection and improvement of task performance when label noise is high.
\end{enumerate}

Overall, the cluster analysis reveals a number of interrelationships between the noise filtering properties of the methods and the resulting downstream performance. At the same time, it becomes clear that the framework conditions (noise type, noise level, and data set) must be taken into account in order to explain the behavior of the methods. Establishing simple, generalizable linear correlations is therefore only possible to a very limited extent. In line with this, the synergetic analysis once again illustrates that the analyzed data-centric methods may behave differently in some of the potential metrics of interest. Good label noise filter performance, accurate noise level estimation, and improvement in task performance are only partially linked.

\section{Conclusions}\label{sec:Conclusions}

The systematic evaluation of models trained under injected label noise shows that data-centric methods can be highly relevant for remote sensing data. Even for low noise levels of 10\%, which can easily occur in real data sets, our analyses show the detrimental effect of label noise. The use of data-centric methods leads to systematic and drastic improvements in the generalization ability of the models. This applies to all of the methods tested, even at medium and higher noise levels up to and including 50\%. Beyond this potential, our analyses demonstrated further advantages of data-centric methods in terms of the ability to specifically identify the erroneous labels. Here, the methods examined prove to be equally capable of analyzing label noise properties of existing noisy data sets and providing cleaned data sets. The decomposition of the performance of data-centric methods into these two main parts - label flaw identification properties and influence on scene classification task accuracies - not only enables us to understand the behavior of the methods, but also to derive recommendations for the choice of a specific data-centric method. It is apparent that depending on the anticipated factors (noise level and noise type) and the ultimate goal (noise identification, noise level estimation or task performance), different methods may be favorable. While \ac{TopoFilter} seems particularly suitable for label noise identification, \ac{CL} reveals its strengths primarily in noise level estimation. If the focus is on the final task accuracies, the choice of method should be made dependent on the specific conditions (noise level and type). The correlations we analyzed between label noise identification and task performance are also relevant as they show that it is insufficient to conclude about the general performances of a label noise handling method simply based on assessing one metric.

Our analyses also reveal directions for further research. The partial data set-specific variability of the results makes it difficult to generalize findings about the behavior of the methods to other data sets. The need to test data-centric methods on other data sets is underscored by the limited nature of the data sets examined here. Real \ac{RS} data sets are usually inherently more difficult to classify and have characteristics such as more pronounced class imbalance. Our analyses also show that more realistic assumptions about the noise processes, using asymmetric instead of simple uniform models, limit the performance of the analyzed methods. Accordingly, extending the analyses to include other realistic assumptions, such as feature-/instance-dependent noises, would also be informative in order to better assess the performance of data-centric methods in realistic \ac{RS} scenarios. Our work provides a suitable basis for such future analyses.

\section{Acknowledgements}

We would like to thank Timo Stomberg and the anonymous reviewers for their helpful input in improving the quality of this manuscript. This work has been partially funded by the Federal Ministry of Transport (BMV) as part of the mFUND innovation initiative, project name: KIBI, funding code: 19F2276.

{
	\begin{spacing}{1.17}
		\normalsize
		\bibliography{references} 
	\end{spacing}
}

\end{document}